\documentclass{ifacconf}

\makeatletter
\let\old@ssect\@ssect % Store how ifacconf defines \@ssect
\makeatother

\counterwithin*{section}{part}

\usepackage{graphicx}      % include this line if your document contains figures
\usepackage{natbib}        % required for bibliography
\usepackage{amsmath,amssymb,amsfonts,bbm}

\usepackage{cancel}
\usepackage{float}
\usepackage{subcaption}
\usepackage{xcolor}
\usepackage{multirow}
\usepackage[colorlinks,allcolors=blue,draft=false]{hyperref}
\usepackage[capitalise]{cleveref}
\crefname{equation}{}{}
\usepackage{siunitx}
\usepackage{algorithmicx,algorithm,algpseudocodex}
\usepackage{blindtext} 
\DeclareMathOperator*{\argmin}{arg\,min}

\makeatletter
\def\@ssect#1#2#3#4#5#6{%
	\NR@gettitle{#6}% Insert key \nameref title grab
	\old@ssect{#1}{#2}{#3}{#4}{#5}{#6}% Restore ifacconf's \@ssect
}
\makeatother

\begin{document}
% \pagecolor{black}
\begin{frontmatter}
% \color{yellow}

\title{Systematic Evaluation of Trade-Offs in Motion Planning Algorithms for Optimal Industrial Robotic Work Cell Design\thanksref{footnoteinfo}}

\author[First]{Guillaume de Mathelin} 
\author[First]{Christian Hartl-Nesic} 
\author[First,Second]{Andreas Kugi}

\address[First]{Automation \& Control Institute, TU Wien, Vienna, Austria}
\address[Second]{AIT Austrian Institute of Technology GmbH, Vienna, Austria}

\thanks[footnoteinfo]{© 2025 the authors. This work has been accepted to IFAC for publication under a Creative Commons Licence CC-BY-NC-ND.}

\begin{abstract}                % Abstract of not more than 250 words.
	The performance of industrial robotic work cells depends on optimizing various hyperparameters referring to the cell layout, such as robot base placement, tool placement, and kinematic design.
	Achieving this requires a bilevel optimization approach, where the high-level optimization adjusts these hyperparameters, and the low-level optimization computes robot motions.
	However, computing the optimal robot motion is computationally infeasible, introducing trade-offs in motion planning to make the problem tractable.
	These trade-offs significantly impact the overall performance of the bilevel optimization, but their effects still need to be systematically evaluated.
	In this paper, we introduce metrics to assess these trade-offs regarding optimality, time gain, robustness, and consistency.
	Through extensive simulation studies, we investigate how simplifications in motion-level optimization affect the high-level optimization outcomes, balancing computational complexity with solution quality.
	The proposed algorithms are applied to find the time-optimal kinematic design for a modular robot in two palletization scenarios.

\end{abstract}


\end{frontmatter}
\section{INTRODUCTION}

The performance of industrial robotic work cells crucially determines the efficiency and throughput in modern manufacturing.
The most important performance indicators are the cycle time, the energy consumption per cycle, and total cost, which are significantly influenced by, e.g., robot base placement \citep{balci2023optimal}, tool placement \citep{weingartshofer2021optimal}, kinematic design of the robot \citep{kulz2024optimizing}, work cell layout \citep{balci2023optimal}, and grasping poses \citep{zimmermann2020multi}.
In order to adjust and optimize robotic work cells with respect to certain factors, called \emph{hyperparameters}, the holistic process scenario needs to be considered \citep{wachter2024robot}, i.e., including the (collision) environment, the task sequence, and the robot motions.
Obtaining the robot motions for a certain task sequence in a given environment is a time-consuming optimization problem by itself, as the kinematic and dynamic limits of the robot need to be respected systematically.
Hence, optimizing the hyperparameters in a robot work cell inherently requires planning and optimizing the process that solves the task at a lower level.
The interaction between the two optimization problems is known in the literature as a \emph{bilevel optimization structure} \citep{zimmermann2020multi, tang2019time}.

\begin{figure}[!t]
	\includegraphics[width=0.47\textwidth]{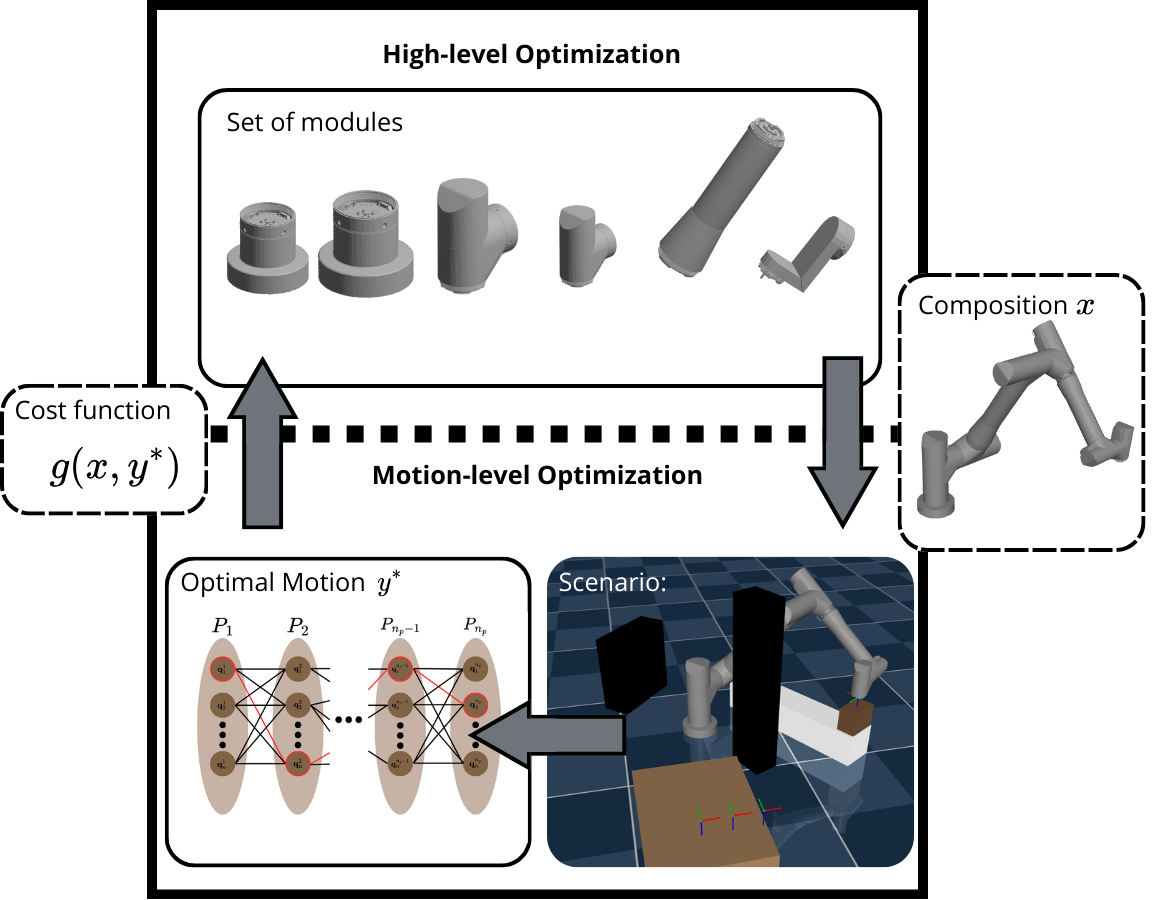}
	\caption{An example of a bilevel optimization structure: The high-level algorithm gives compositions to the lower level, which returns the corresponding optimal cost values.}
\label{intro:bilevel}
\end{figure}

Finding the optimal kinematic design for a modular robot in a given pick-and-place scenario is an example of a bilevel optimization, see \cref{intro:bilevel}. Here, the \emph{high-level} optimization determines the module composition, and the \emph{low-level} optimization solves the complete robot motion for a given composition.
In order to solve this task, the robot generally passes a sequence of poses, i.e., end-effector positions and orientations.
To this end, the low-level optimization has to compute robot trajectories in joint space to move between each pair of consecutive poses.
In this work, this is referred to as \emph{motion-level optimization}.

Several challenges arise when using motion-level optimization in a bilevel structure.
Firstly, different motion constraints, such as the kinematics and dynamics of the robot and collision avoidance, make the motion-level optimization computationally expensive.
Secondly, each task pose is reachable with multiple joint configurations requiring to consider many different paths between the poses.
Thirdly, kinematically redundant robots \citep{franka2016} and robots with infinite rotation \citep{beckhoff2022attro,robco} become widely available on the market.
These robots even have infinite joint-space solutions for each task pose.
Thus, an exhaustive search considering the motions between all joint configurations is computationally infeasible, and the globally optimal solution cannot be obtained within the motion-level optimization.

Consequently, many works resort to suboptimal solutions by introducing compromises and \emph{trade-offs} to reduce the computation time of motion-level optimization.
For instance, heuristic approaches have been applied to handle the multiplicity of joint configurations for task poses \citep{whitman2018task}.
In this way, only the most promising joint configurations are selected for trajectory planning between the sequence of task poses.
In order to consider the complete robot dynamics and the (collision) environment, \cite{wachter2024robot} propose to use a global optimization approach that generates a near-optimal and collision-free trajectory between task poses, e.g., using VP-STO \citep{jankowski2023vp} or STOMP \citep{kalakrishnan2011stomp}.
Other approaches use local motion optimization strategies to transition between consecutive task poses \citep{icer2016cost, zimmermann2020multi}.
Sampling-based methods \citep{sucan2012open} in combination with trajectory optimizers \citep{kunz2013time,pham2018new} have also been implemented in the motion-level optimization \citep{kulz2024optimizing, icer2017evolutionary, liu2020optimizing}.
Finally, the motion-level planning problem was also simplified in some works by ignoring the collision constraints \citep{icer2016task} or by neglecting the optimality of the motion \citep{ha2018computational}.
Note that none of the above approaches have yet considered robots with infinite rotations, which is often a beneficial property for obtaining optimal robot motions. 

Simplifying the motion-level optimization reduces the computational complexity, but it may also impact the results and quality of the high-level optimization.
This introduces trade-offs between computation time and the quality of the obtained solution.
As the high-level optimization directly depends on the output of the motion-level optimization, the introduced compromises immediately impact the high-level process.
For instance, motion-level planning may yield solutions that are far from optimal and guide the high-level optimization to a suboptimal solution.
As many motion planners comprise stochastic elements, e.g., sampling-based methods like RRT-Connect or VP-STO, their output may vary significantly for repeated runs with the same input parameters, and potentially ruin the convergence of the high-level optimization.

Therefore, this work aims to systematically evaluate trade-offs in motion-level optimization and investigate the impacts on high-level optimization.
The main contributions of this paper are as follows:
\begin{itemize}
    \item the introduction of metrics for optimality, computation time gain, consistency, and robustness to evaluate and compare different trade-offs in the motion-level optimization,
    \item the consideration of robots with infinite rotations,
    \item the evaluation of these metrics using simulation studies and the application to a bilevel optimization structure.
\end{itemize}

\section{Metrics}
\label{sec:metrics}
A bilevel optimization structure is mathematically formulated as
\begin{equation}
\label{eq:bilevel-optimization}
x^{*} = \argmin_{x \in \mathcal{C}_x} g^*(x, \theta)
\end{equation}
with
\begin{equation}
\label{eq:motion_equation}
    \quad g^*(x, \theta) = \min_{y \in \mathcal{C}_y} f(x, y, \theta)~,
\end{equation}
where the outer (high-level) optimization finds the optimal hyperparameters $x$ from a set $\mathcal C_x$ using the inner (motion-level) optimization, which computes the optimal robot trajectory $y$ from the set of all feasible robot motions $\mathcal C_y$.
The motion-level optimization \cref{eq:motion_equation} uses a cost function $f(x,y,\theta)$ with the set of fixed parameters $\theta$ containing, e.g., the sequence of task poses and a description of the environment.
For the high-level optimization \cref{eq:bilevel-optimization}, the cost function $g^{*}(x,\theta)$ refers to the optimal motion-level algorithm without trade-offs, and the suboptimal cost function $\hat g(x,\theta)$ describes an algorithm that uses heuristics to reduce the computational complexity, called $\emph{variant}$.
Four metrics are introduced in the following to evaluate and quantify these variants.
These metrics are formulated using statistical functions, as many motion-level planners comprise stochastic elements.

Firstly, the \emph{optimality score} $O(\hat g)$ of a variant $\hat g$ indicates how far its solutions deviate from the optimal algorithm $g^*$.
This criterion is given by the expectation values $E(\cdot)$ of the relative deviation from the optimum as
\begin{equation}\label{optimality}
    O(\hat{g}) = E\left( \frac{\hat{g}(x, \theta)-g^*(x, \theta)}{g^*(x, \theta)}\right)~.
\end{equation}

Secondly, the \emph{time gain score} $T(\hat g)$ evaluates the computation time gain.
It is obtained as the expectation value of the relative gain between the computation time $t^*$ of the optimal algorithm $g^*$ and the computation time $\hat t$ of the variant $\hat g$ using
\begin{equation}\label{time_eq}
    T(\hat{g}) = E\left( \frac{t^*(x, \theta) - \hat{t}(x, \theta)}{t^*(x, \theta)}\right)~.
\end{equation}
As the trade-offs are expected to reduce the computation time, $T(\hat g)\geq0$ holds.

Thirdly, the \emph{robustness score} $R(\hat g)$ evaluates how reliable a variant $\hat g$ is in finding a solution if one exists.
This score is of interest as a variant $\hat g$ might omit a significant portion of valid solutions if it uses excessive simplifications in the algorithm.
This robustness score $R(\hat g)$ is quantified as the probability $\mathbb P(\cdot)$ of the variant $\hat g$ to find a solution, i.e., $\hat{g}(x,\theta) \neq \infty$, when also the optimal algorithm $g^*$ does.
Mathematically, this is given by
\begin{equation}\label{robust}
    R(\hat{g}) = \mathbb{P}\big(\hat{g}(x,\theta) \neq \infty \:|\: g^*(x,\theta) \neq \infty\big)~.
\end{equation}
 
Finally, the \emph{consistency score} $C(\hat g)$ indicates how much the output of $\hat g(x,\theta)$ varies when repeatedly run with the same inputs $x$ and $\theta$.
It is computed using the relative variance of the output as
\begin{equation}\label{consistent}
    C(\hat{g}) =\frac{V(\hat{g}(x, \theta))}{E(\hat{g}(x, \theta))}~,
\end{equation}
with the variance $V(\cdot)$.
\section{Motion-Level Optimization}
\label{sec:motion-level-optimization}
This section summarizes four different algorithms for motion-level optimization, namely the (near-)optimal algorithm $g^*$, a variant using the $\textrm A^*$ algorithm \citep{de2007geospatial}, one using a two-stage VP-STO (referred to as Spline variant), and a last one using OMPL \citep{sucan2012open}.
All algorithms compute the joint motion and the cost function value $g$ for moving through a given sequence of $n_\mathrm p$ task poses $P=[\mathbf p_1,\dotsc,\mathbf p_{n_\mathrm p}]$.
In our case, the cost function value $g$ denotes the time duration of the motion.
All programs are implemented in C++ and use Pinocchio \citep{carpentier2019pinocchio} and MuJoCo \citep{todorov2012mujoco} to simulate the robots and environments.

\subsection{Near-Optimal Algorithm}
Finding the global optimal robot motion is computationally expensive, even for a standard 6-axis robot and a short sequence of task poses.
Hence, such an approach is not suitable for simulation studies with a large number of samples.
Therefore, this work uses the near-optimal algorithm proposed by \cite{wachter2024robot} as a basis for the optimal algorithm $g^*$, listed in \cref{alg:algo_optimal} as pseudo-code.
This algorithm first constructs a graph $\mathcal{G}$ (composed of nodes $\mathcal{V}$ and edges $\mathcal{E}$) of all joint motions between the task poses $P$ and then applies Dijkstra's algorithm \citep{de2007geospatial} to find the complete optimal robot motion.
\begin{algorithm}[tb]
	\caption{Near-Optimal Algorithm and $\textrm{A}^*$ Variant}
	\label{alg:algo_optimal}
	\begin{algorithmic}[1]
		\Procedure{GraphExploration}{$P$}
		\State $\mathcal{V} \leftarrow \{\}$; $\mathcal{E} \leftarrow \{\}$; $Q \leftarrow \{\};$
		\For{$i = 1$ \textbf{to} $n_\mathrm p$}
		\State $\textbf{p}_{i} \leftarrow P[i]$
		\State $Q[i] \leftarrow \texttt{InverseKinematics}(\textbf{p}_i)$
		\State $\mathcal{V} \leftarrow \mathcal{V} \cup Q[i]$
		\If{$i \neq 1$}
		\For{$\textbf{q}_{\mathrm{s}}\in Q[i-1]$ \textbf{and} $\textbf{q}_{\mathrm{e}} \in Q[i]$}
		\State $\mathcal{E} \leftarrow \mathcal{E} \cup \{[\textbf{q}_{\mathrm{s}},\textbf{q}_{\mathrm{e}}]\}$
		\EndFor
		\EndIf
		\EndFor
		\State $\mathcal{G} \leftarrow (\mathcal{V},\mathcal{E})$
		\State $g \leftarrow \texttt{Exploration}(\mathcal{G} )$
		\State \textbf{return} $g$
		\EndProcedure
	\end{algorithmic}
\end{algorithm}
\subsubsection{Graph Building}
For each task pose $\textbf{p}_{i}$, $i=1,\dotsc,n_\mathrm p$, the near-optimal algorithm starts with the computation of its corresponding inverse kinematics solutions and saves them as a list in  $Q[i]$ (lines 4--6).
These solutions form the nodes of the graph.
Then, the edges are created between all inverse kinematics solutions $Q[i]$ and the previous ones $Q[i-1]$ (lines 8--9).

For a robot with $n_\mathrm q$ joints, the inverse kinematics solver \texttt{InverseKinematics} returns joint configurations in the range $\pm\pi$ for each generalized coordinate in $\textbf q=[q_1,q_2,\dotsc,q_{n_\mathrm q}]$.
The inverse kinematics solver is a Jacobian-based numerical solver, which is executed multiple times with different initial configurations to ensure coverage of the joint space.

\subsubsection{Infinite Rotation}
The infinite rotation property means that one or more joint axes of the robot can rotate infinitely.
This implies an infinite number of joint-space solutions for each task pose and, consequently, an infinite number of possible joint motions between two consecutive task poses.
In order to make this circumstance computationally tangible, we assume that each joint may move up to $2\pi$ during a single joint motion between two task poses.
Due to the infinite rotation property, each joint has two ways to reach the end configuration, i.e., moving in positive or negative direction.
For a robot with $n_\mathrm q$ joints, each joint can individually move either in the positive or negative direction, and all combinations must be considered.
Consequently, for a single edge between two joint configurations, there are $2^{n_\mathrm{q}}$ possible ways.
\subsubsection{Edge Evaluation}
In \texttt{Exploration} (line 11), Dijkstra's algorithm explores the graph, evaluates progressively the cost function values of the edges and returns the cost function value $g$.
For evaluating the cost value of a single edge, as there are $2^{n_\mathrm{q}}$ ways, a valid trajectory for each way is computed with the VP-STO algorithm and only the optimal one is selected.
For finding a valid trajectory, the VP-STO algorithm optimizes the global cost function $c$ chosen as
\begin{equation}\label{VPSTO_accurate}
	c = T + 100\,n_\mathrm c~,
\end{equation}
where $T$ is the time duration of the trajectory, and $n_\mathrm c$ is the number of trajectory points that violate a constraint limit (joint torque limits or collision contacts).

\subsection{\texorpdfstring{$\textrm{A}^*$}{A*}  Variant}
The $\textrm A^*$ variant is a motion-level algorithm that uses the heuristic $\textrm A^*$ algorithm to speed up the motion planning and evaluate the cost function.
The structure of the algorithm is identical to the near-optimal algorithm, see \cref{alg:algo_optimal}, but the function \texttt{Exploration} (line 11) uses the $\textrm A^*$ algorithm for the graph search.
The heuristic distance estimate for one edge $(\textbf{q}_{\mathrm{s}},\textbf{q}_{\mathrm{e}})$ is chosen as
\begin{equation}
    h = \max_{i=1,...,n_\mathrm{q}}{\frac{|q_{\mathrm{s},i} - q_{\mathrm{e},i}|}{\dot{q}_{\mathrm{max},i}}}~,
    \label{eq:astar_heuristic}
\end{equation}
with the joint velocity limit $\dot{q}_{\mathrm{max},i}$ of axis $i$.
The estimate in \cref{eq:astar_heuristic} is the trajectory duration if the robot would move at its joint velocity limit for the slowest axis.
Due to the infinite rotation property, $2^{n_\mathrm q}$ heuristic values \cref{eq:astar_heuristic} are computed for the edge and only the minimal one is kept.

\subsection{Spline Variant}
\begin{algorithm}[t]
	\caption{Spline and OMPL-10 Variant}
	\label{alg:algo_vpsto_variant}
	\begin{algorithmic}[1]
		\Procedure{GreedyVariant}{$P$}
		\State $g \leftarrow 0$;
		\State $\textbf{Q}_{\mathrm{e}} \leftarrow \texttt{InverseKinematics}(P[1])$
		\State $\textbf{q}_{\mathrm{s}} \leftarrow \texttt{MinimalNorm}(\textbf{Q}_{\mathrm{e}})$
		\For{$i = 2$ \textbf{to} $n_\mathrm p$}
		\State $\textbf{p}_{i} \leftarrow P[i]$
		\State $\textbf{Q}_\mathrm{e} \leftarrow \texttt{InverseKinematics}(\textbf{p}_i)$
		\State $\tilde{\textbf{Q}}_{\mathrm{e}} = \{\};$
		\For{$\textbf{q}_{\mathrm{e}} \in \textbf{Q}_\mathrm{e}$}
		\State $\textbf{Q} \leftarrow \texttt{Extend}(\textbf{q}_{\mathrm{s}},\textbf{q}_{\mathrm{e}})$
		\State $\tilde{\textbf{Q}}_{\mathrm{e}} \leftarrow \tilde{\textbf{Q}}_{\mathrm{e}} \cup \textbf{Q}$
		\EndFor
		\State $\tilde{\textbf{Q}}_{\mathrm{e}} \leftarrow \texttt{SortDistance}(\tilde{\textbf{Q}}_{\mathrm{e}},\textbf{q}_{\mathrm{s}})$
		\State $\textrm{SolutionFound} \leftarrow \textrm{False}$
		\For{$\tilde{\textbf{q}}_{\mathrm{e}} \in \tilde{\textbf{Q}}_{\mathrm{e}}$}
		\State $g_{\mathrm{e}} \leftarrow \texttt{Connection}(\textbf{q}_{\mathrm{s}},\tilde{\textbf{q}}_{\mathrm{e}})$
		\If{$g_{\mathrm{e}} \neq +\infty$}
		\State $\textrm{SolutionFound} \leftarrow \textrm{True}$
		\State $g \leftarrow g+g_{\mathrm{e}}$
		\State $\textbf{q}_{\mathrm{s}} \leftarrow \tilde{\textbf{q}}_{\mathrm{e}}$
		\State \textbf{break}
		\EndIf
		\EndFor
		\If{\textbf{not} $\textrm{SolutionFound}$}
		\State \textbf{return} $+\infty$
		\EndIf
		\EndFor
		\State \textbf{return} $g$
		\EndProcedure
	\end{algorithmic}
\end{algorithm}
The Spline variant significantly reduces the computational complexity by following a greedy approach, see \cref{alg:algo_vpsto_variant}.
After computing the inverse kinematics solutions $\mathbf Q_\mathrm e$ of the first task pose $P[1]$, the variant selects the solution with the minimal Euclidean norm as the start configuration $\mathbf q_\mathrm s$ for the joint motion (lines 3--4).
Then, the Spline variant efficiently computes the joint motions for the remaining task poses $\textbf{p}_{i}$, $i=2,\dotsc,n_\mathrm p$ (lines 5--22) as follows.
First, all inverse kinematics solutions $\textbf{Q}_\mathrm{e}$ for $\textbf{p}_{i}$ are computed (line 7).
Then, as there are $2^{n_\mathrm{q}}$ ways between the start configuration $\textbf{q}_{\mathrm{s}}$ and any inverse kinematics solution $\textbf{q}_{\mathrm{e}} \in \textbf{Q}_\mathrm{e}$, their ending joint position values $\textbf{Q}$ are computed in $\texttt{Extend}$ and saved in $\tilde{\textbf{Q}}_{\mathrm{e}}$ (lines 9--11).
Once all joint position values are computed and gathered in $\tilde{\textbf{Q}}_{\mathrm{e}}$, they are sorted according to their distance to $\textbf{q}_{\mathrm{s}}$ (line 12).
Then, the algorithm sequentially computes joint motions to the candidate joint position values $\tilde{\textbf{q}}_{\mathrm{e}} \in \tilde{\textbf{Q}}_{\mathrm{e}}$ and stops when the planning is successful (lines 13--20).
Once a joint-space motion for reaching $\textbf{p}_{i}$ is found, its corresponding cost $g_{\mathrm{e}}$ is added to the cost function value $g$ (line 18) and its associate joint position value is selected as the new start configuration $\mathbf q_\mathrm s$ for the next task pose (line 19).
The joint motion between $\textbf{q}_{\mathrm{s}}$ and any joint position value $\tilde{\textbf{q}}_{\mathrm{e}} \in \tilde{\textbf{Q}}_{\mathrm{e}}$ is computed in \texttt{Connection} with VP-STO running twice (line 15).
First, in order to further reduce computation time, VP-STO is run with a limited number of iterations to check if a collision-free motion in joint space exists.
Second, if the check is positive, VP-STO is called again with a high number of iterations and a large population size to obtain the time-optimal robot trajectory and the associated cost function value $g_{\mathrm{e}}$.

\subsection{OMPL-10 Variant}
The OMPL-10 variant also uses the greedy approach of the previous variant and only differs in the implementation of the function \texttt{Connection} (line 15 in \cref{alg:algo_vpsto_variant}).
This variant combines the AIT* sampling-based planner \citep{strub2020adaptively} with the time-optimal trajectory algorithm TOPP-RA \citep{pham2018new} to compute the joint-space motions.
The sampling-based planner is implemented using the OMPL library and has a computation time limit of \qty[mode = text]{10}{\second} for finding a path (hence the name of the variant).
If the motion is not found within the given time span, the respective ending joint position is omitted.

\section{Use Case: Time-Optimal Kinematic Design for Palletization Tasks}
This section discusses the metrics from \cref{sec:metrics} applied to the motion-level planning algorithms described in \cref{sec:motion-level-optimization} for a palletization scenario.
This scenario aims to find the time-optimal kinematic design for a modular robot composed of modules with infinite rotation \citep{beckhoff2022attro}. To this end, a bilevel optimization structure is used, see \cref{intro:bilevel}. The high-level optimization searches for the time-optimal robot composition and uses the motion-level optimization to compute the robot motions and evaluate each design.

First, the variables $x$ and $\theta$ for the cost functions $g^*(x,\theta)$ and $\hat g(x,\theta)$ are described in detail.
Second, the metrics for optimality, time gain, robustness, and consistency are evaluated for the near-optimal algorithm and the three variants of \cref{sec:motion-level-optimization}, using an input population $(x,\theta)$ with large variation.
Finally, these variants for the motion-level planners are utilized in the bilevel optimization problem for two palletization environments. A video presenting the different results and simulations is available at \href{https://www.acin.tuwien.ac.at/1e06/}{acin.tuwien.ac.at/1e06}.
\subsection{Palletization Scenario}
\begin{figure}
	\centering
	\begin{subfigure}[b]{0.23\textwidth}
		\includegraphics[width=\textwidth]{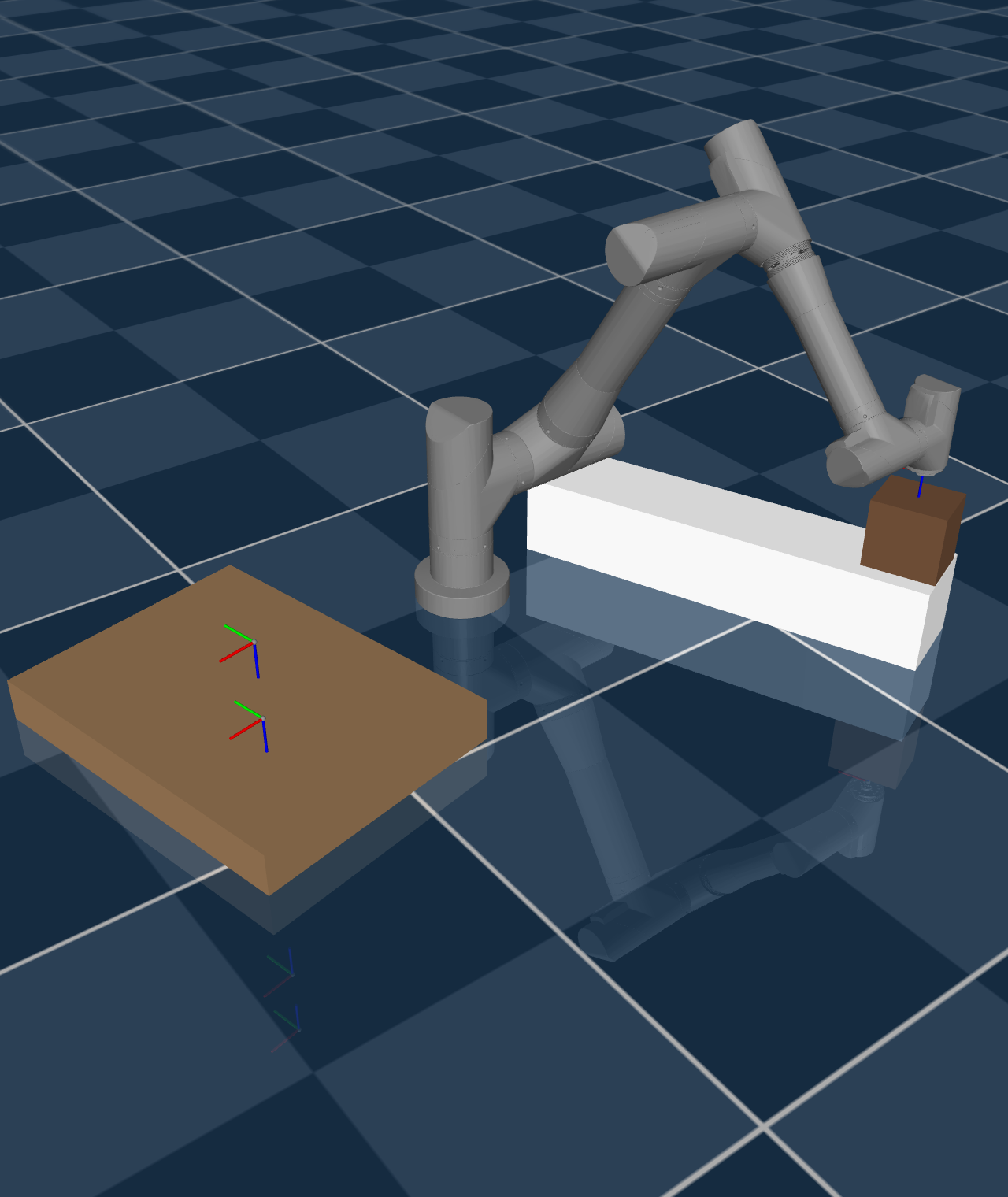} % Replace with your image file
	\end{subfigure}
	\begin{subfigure}[b]{0.23\textwidth}
		\includegraphics[width=\textwidth]{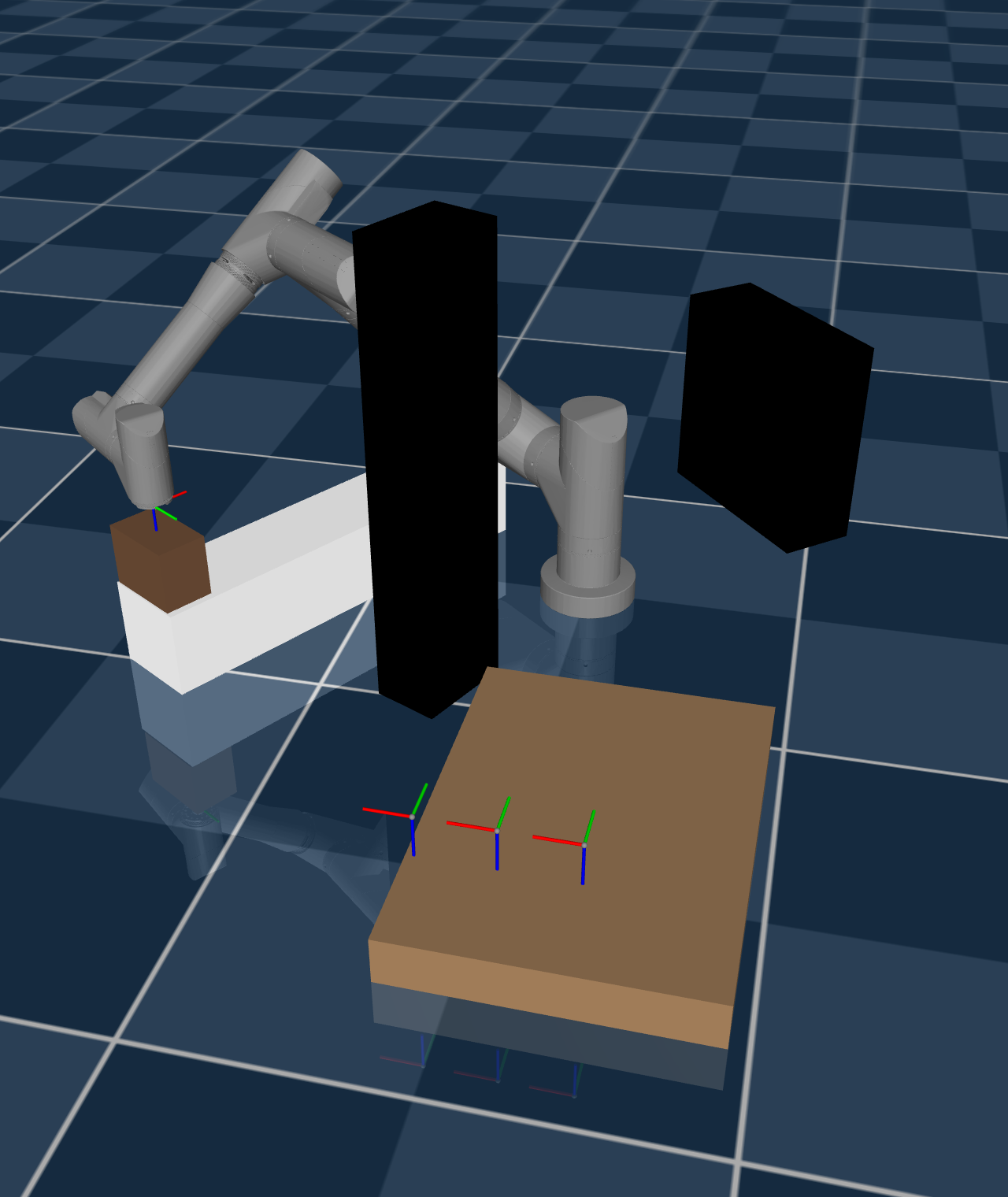} % Replace with your image file
	\end{subfigure}
	\caption{Random scenario generator for palletization environments: A simple environment (left) is free of obstacles, while pillars are placed close to the base position in the more complex environment (right).}
	\label{fig:pal_env}
\end{figure}

In this work, a random scenario generator is used to generate different palletization environments and tasks, see two examples in \cref{fig:pal_env}.
Each environment consists of a robot (placed in the origin of the world frame), a conveyor belt (white box), a pallet (light brown box), multiple obstacles (black boxes), and the packet to be palletized (dark brown box).
The task of the robot is to sequentially pick up boxes from the conveyor belt and place them in the predefined locations on the pallet (coordinate frames).
After placing a box on the pallet, the next one appears on the conveyor belt, which results in a sequence of pick-and-place motions that the robot has to perform.
Note that the number of boxes is limited to four to achieve reasonable computation times for the near-optimal algorithm with the infinite rotation property.
In order to focus only on motion planning for the sequence of task poses, the box grasping and releasing motions are not considered inside the motion-level optimization.
The placements of all assets and the number, size, and weight of the boxes are chosen randomly and contained in the fixed parameters $\theta$.
The hyperparameters $x$ describe a modular robot composition, i.e., the specific arrangement of modules to construct the considered modular robot.

\subsection{Metrics Evaluation}
The metrics introduced in \cref{sec:metrics} are evaluated in two experiments in this section.
In the first experiment, the time gain, optimality, and robustness are evaluated for all motion-level algorithms summarized in \cref{sec:motion-level-optimization}.
The second experiment evaluates the consistency of the variants with trade-offs.

These experiments use two types of environments (see \cref{fig:pal_env}), i.e., a simple and a more complex environment.
The simple environment constitutes an obstacle-free scenario.
When pillars and walls close to the robot base obstruct the robot's movement, the environment is more complex.
The set of hyperparameters $C_x$ in \cref{eq:bilevel-optimization} contains various robots with 5 and 6 DoF (degree of freedom).
The input population $(x,\theta)$ comprises 40 different environments (20 of each type) with 5-DoF and 6-DoF robot compositions.
Hence, a large population with distinct variations ensures that the computed metrics are of broader relevance.

\subsubsection{Evaluation of Time Gain, Optimality, and Robustness}
In the first experiment, 250 different input combinations $(x,\theta)$ are executed once for each motion-level algorithm, for which the output results and the computation times are recorded.
The metrics \cref{optimality,time_eq,robust} are computed and summarized in \cref{table:torc_table}.
\begin{table}
	\centering
	\caption{Metrics evaluation for the motion-level optimization algorithms}
	\label{table:torc_table}
	\begin{tabular}{c|ccc}
		\hline
		Variant      & $\textrm{A}^*$ & Spline         & OMPL-10        \\ \hline
		Time gain    & 0.161         & 0.998 & \textbf{0.999} \\
		Optimality   & \textbf{0.00}  & 0.17           & 0.30           \\ 
		Robustness   & \textbf{0.996}  & 0.952  & 0.968  \\
		Consistency  & n/a            & \textbf{0.020} & 0.159          \\ \hline
	\end{tabular}
\end{table}
The scores obtained for the $\mathrm A^*$ variant show that optimality and robustness are practically equivalent to the near-optimal algorithm at the expense of long computation times.
The $\mathrm A^*$ variant only saves approximately \SI{16}{\percent} of computation time because the exploration using the $\mathrm A^*$ algorithm also visits almost all nodes as Dijkstra's algorithm.
In contrast, the greedy variants Spline and OMPL-10 significantly reduce the number of visited nodes, saving \SI{99}{\percent} computation time.
Still, the optimality of the Spline variant degrades only by \SI{17}{\percent} and \SI{30}{\percent} for the OMPL-10 variant.
The robustness score is high for all variants, as a motion-level solution was found for more than \SI{95}{\percent} of the samples.

\subsubsection{Evaluation of Consistency}
In order to evaluate the consistency of the motion-level algorithms according to \cref{consistent}, 100 different input combinations $(x,\theta)$ are evaluated 20 times each. 
A consistency score is calculated for each sample, and the mean scores are reported in \cref{table:torc_table}.
Due to the extensive computation time, the consistency score of the $\mathrm A^*$ variant could not be obtained (n/a).
As this variant behaves very similarly to the near-optimal algorithm in the first experiment, the variance of the cost function is expected to be very low, yielding a consistency score close to zero.
The solutions of the Spline variant show a narrow variation of only \SI{2}{\percent} of the mean, while the OMPL-10 variant exhibits a much wider variation of \SI{16}{\percent}.

\subsubsection{Conclusion}
Regarding time gain, robustness, and consistency metrics, the Spline variant is the best choice for motion-level optimization in a bilevel optimization structure.
While the solutions obtained with this variant are suboptimal by \SI{17}{\percent} on average, the heuristic trade-offs are justified due to the significant time gain.
Therefore, the Spline variant is suitable for the application in the palletization task.

\subsection{Bilevel Optimization using Hierarchical Elimination}
A bilevel optimization structure is employed to find the time-optimal kinematic design of a modular robot for a simple and a more complex palletization scenario, see \cref{fig:pal_env}.
Hierarchical elimination \citep{icer2016cost,icer2016task} is used as high-level optimization, which employs the two greedy variants Spline and OMPL-10 as motion-level optimization.
In the hierarchical elimination, all possible robot compositions are generated and then discarded one by one using a series of criteria with increasing computational complexity.
First, \SI{3.3e6}{} modular robot compositions are generated using up to 10 modules and 6 motors.
After discarding compositions with less than four motors and ones which cannot reach all task poses, or cannot carry the box at the task poses, the motion-level optimization is executed for each of the remaining compositions $x$.
Finally, the high-level optimization returns the optimal composition.
A multi-core processor with 64 cores is used to speed up the computations.

The results in \cref{table:ha_table} indicate that regarding trajectory duration, the Spline variant found better compositions than the OMPL-10 variant in both scenarios.
\begin{table}
	\centering
	\caption{Optimal solution found, mean of the computation time and consistency score for each variant in each environment}
	\label{table:ha_table}
	\begin{tabular}{c|c|cc}
		\hline
		& Variant & Simple       &  Complex       \\ \hline
		Trajectory duration of& Spline & \textbf{1.61}  &  \textbf{3.18}       \\
		  the optimal composition (\unit{\second})& OMPL-10&1.81 & 3.42       \\  \hline
		Computation time (\unit{\hour}) & Spline& 7.1  &   \textbf{12.8}         \\
		(64 cores) & OMPL-10& \textbf{4.0}   &   15.6           \\ \hline
		\multirow{2}{*}{Consistency score } & Spline& 0.031  &   \textbf{0.044}         \\
		& OMPL-10& \textbf{0.0001}   &   0.296          \\ \hline
	\end{tabular}
\end{table}
The multi-core computation time is lower for the OMPL-10 variant in the simple scenario, while in the complex scenario, the Spline variant is evaluated faster.
The main reason is that the Spline variant takes less time to identify an invalid composition.
As more compositions are invalid for the complex environment, the Spline variant evaluates the complex scenario faster.

Next, the consistency of the cost function is checked for the 100 best compositions found using hierarchical elimination.
Each robot composition is evaluated 20 times.
In terms of consistency, the OMPL-10 variant performs significantly better in the simple environment, while the Spline variant performs equivalently in both scenarios.
The main reason lies in the construction of the OMPL-10, which selects the direct joint-space solutions when available. This condition is predominantly met in the simple environment.
Finally, for the time-optimal composition for palletization task problems, the Spline variant is an appropriate motion-level optimizer and can replace the near-optimal algorithm, which would take more than $100$ days with $64$ cores to return its solutions.

\section{Conclusion}
In a bilevel optimization structure, the high-level optimization directly depends on the output quality of the motion-level algorithm.
Any introduced trade-offs have a direct impact on the found optimum.
This work introduces four novel metrics, namely optimality, time gain, robustness, and consistency, to evaluate the motion-level algorithm.
These metrics allow us to measure the trade-offs between computation time and quality of the result in a systematic and quantitative way.

For motion-level optimization, a near-optimal algorithm and three variants with trade-offs, i.e., the $\mathrm A^*$, the Spline, and the OMPL-10 variant, are proposed.
A large simulation study using a scenario generator for palletization tasks was conducted.
This study shows that the Spline variant performs close to the near-optimal algorithm in optimality, robustness, and consistency while significantly reducing the computational complexity.
Finally, the Spline and the OMPL-10 variants are used to efficiently find the time-optimal modular robot composition for a given palletization scenario, where the Spline variant was shown to be more suitable for the bilevel optimization problem. 

Future work will broaden the simulation study to a larger variation of scenarios and optimization goals, e.g., minimizing energy consumption or wear and tear.

%%%%%%%%%%%%%%%%%%%%%%%%%%%%%%%%%%%%%%%%%%%%%%%%%%%%%%%%%%%%%%%%%%%%%%%%%%%%%%%%

\begin{ack}
	The authors gratefully acknowledge the financial support of Beckhoff GmbH \& Co.\ KG.
\end{ack}

\bibliography{paper} % Entries are in the refs.bib file

\begin{thebibliography}{24}
\providecommand{\natexlab}[1]{#1}
\providecommand{\url}[1]{\texttt{#1}}
\providecommand{\urlprefix}{URL }
\expandafter\ifx\csname urlstyle\endcsname\relax
  \providecommand{\doi}[1]{doi:\discretionary{}{}{}#1}\else
  \providecommand{\doi}{doi:\discretionary{}{}{}\begingroup \urlstyle{rm}\Url}\fi

\bibitem[{Balci et~al.(2023)Balci, Donovan, Roberts, and Corke}]{balci2023optimal}
Balci, B., Donovan, J., Roberts, J., and Corke, P. (2023).
\newblock Optimal workpiece placement based on robot reach, manipulability and joint torques.
\newblock In \emph{Proc. ICRA 2023}, 12302--12308.

\bibitem[{Beckhoff(2022)}]{beckhoff2022attro}
Beckhoff (2022).
\newblock {ATRO}: {Automation} {Technology} for {Robotics}.
\newblock \urlprefix\url{https://www.beckhoff.com/en-en/products/motion/atro-automation-technology-for-robotics/}.

\bibitem[{Carpentier et~al.(2019)Carpentier, Saurel, Buondonno, Mirabel, Lamiraux, Stasse, and Mansard}]{carpentier2019pinocchio}
Carpentier, J., Saurel, G., Buondonno, G., Mirabel, J., Lamiraux, F., Stasse, O., and Mansard, N. (2019).
\newblock The pinocchio {C++} library -- a fast and flexible implementation of rigid body dynamics algorithms and their analytical derivatives.
\newblock In \emph{Proc. SII 2019}, 614--619.

\bibitem[{De~Smith et~al.(2007)De~Smith, Goodchild, and Longley}]{de2007geospatial}
De~Smith, M.J., Goodchild, M.F., and Longley, P. (2007).
\newblock \emph{Geospatial analysis: a comprehensive guide to principles, techniques and software tools}.
\newblock Troubador publishing ltd.

\bibitem[{{Franka Robotics}(2016)}]{franka2016}
{Franka Robotics} (2016).
\newblock {Franka Robotics}.
\newblock \urlprefix\url{https://franka.de}.

\bibitem[{Ha et~al.(2018)Ha, Coros, Alspach, Bern, Kim, and Yamane}]{ha2018computational}
Ha, S., Coros, S., Alspach, A., Bern, J.M., Kim, J., and Yamane, K. (2018).
\newblock Computational design of robotic devices from high-level motion specifications.
\newblock \emph{IEEE Transactions on Robotics}, 34(5), 1240--1251.

\bibitem[{Icer and Althoff(2016)}]{icer2016cost}
Icer, E. and Althoff, M. (2016).
\newblock Cost-optimal composition synthesis for modular robots.
\newblock In \emph{Proc. CCA 2016}, 1408--1413.

\bibitem[{Icer et~al.(2016)Icer, Giusti, and Althoff}]{icer2016task}
Icer, E., Giusti, A., and Althoff, M. (2016).
\newblock A task-driven algorithm for configuration synthesis of modular robots.
\newblock In \emph{Proc. ICRA 2016}, 5203--5209.

\bibitem[{Icer et~al.(2017)Icer, Hassan, El-Ayat, and Althoff}]{icer2017evolutionary}
Icer, E., Hassan, H.A., El-Ayat, K., and Althoff, M. (2017).
\newblock Evolutionary cost-optimal composition synthesis of modular robots considering a given task.
\newblock In \emph{Proc. IROS 2017}, 3562--3568.

\bibitem[{Jankowski et~al.(2023)Jankowski, Bruderm{\"u}ller, Hawes, and Calinon}]{jankowski2023vp}
Jankowski, J., Bruderm{\"u}ller, L., Hawes, N., and Calinon, S. (2023).
\newblock {VP-STO}: Via-point-based stochastic trajectory optimization for reactive robot behavior.
\newblock In \emph{Proc. ICRA 2023}, 10125--10131.

\bibitem[{Kalakrishnan et~al.(2011)Kalakrishnan, Chitta, Theodorou, Pastor, and Schaal}]{kalakrishnan2011stomp}
Kalakrishnan, M., Chitta, S., Theodorou, E., Pastor, P., and Schaal, S. (2011).
\newblock {STOMP}: Stochastic trajectory optimization for motion planning.
\newblock In \emph{Proc. ICRA 2011}, 4569--4574.

\bibitem[{K{\"u}lz and Althoff(2024)}]{kulz2024optimizing}
K{\"u}lz, J. and Althoff, M. (2024).
\newblock Optimizing modular robot composition: A lexicographic genetic algorithm approach.
\newblock In \emph{Proc. ICRA 2024}, 16752--16758.

\bibitem[{Kunz and Stilman(2013)}]{kunz2013time}
Kunz, T. and Stilman, M. (2013).
\newblock Time-optimal trajectory generation for path following with bounded acceleration and velocity.
\newblock \emph{Robotics: Science and Systems VIII}, 209.

\bibitem[{Liu and Althoff(2020)}]{liu2020optimizing}
Liu, S.B. and Althoff, M. (2020).
\newblock Optimizing performance in automation through modular robots.
\newblock In \emph{Proc. ICRA 2020}, 4044--4050.

\bibitem[{Pham and Pham(2018)}]{pham2018new}
Pham, H. and Pham, Q.C. (2018).
\newblock A new approach to time-optimal path parameterization based on reachability analysis.
\newblock \emph{IEEE Transactions on Robotics}, 34(3), 645--659.

\bibitem[{{RobCo}(2020)}]{robco}
{RobCo} (2020).
\newblock {RobCo}, smart modular robots to supercharge your factory.
\newblock \urlprefix\url{https://www.robco.de}.

\bibitem[{Strub and Gammell(2020)}]{strub2020adaptively}
Strub, M.P. and Gammell, J.D. (2020).
\newblock Adaptively informed trees ({AIT*}): Fast asymptotically optimal path planning through adaptive heuristics.
\newblock In \emph{Proc. ICRA 2020}, 3191--3198.

\bibitem[{Sucan et~al.(2012)Sucan, Moll, and Kavraki}]{sucan2012open}
Sucan, I.A., Moll, M., and Kavraki, L.E. (2012).
\newblock The open motion planning library.
\newblock \emph{IEEE Robotics \& Automation Magazine}, 19(4), 72--82.

\bibitem[{Tang et~al.(2019)Tang, Sun, and Hauser}]{tang2019time}
Tang, G., Sun, W., and Hauser, K. (2019).
\newblock Time-optimal trajectory generation for dynamic vehicles: A bilevel optimization approach.
\newblock In \emph{Proc. IROS 2019}, 7644--7650.

\bibitem[{Todorov et~al.(2012)Todorov, Erez, and Tassa}]{todorov2012mujoco}
Todorov, E., Erez, T., and Tassa, Y. (2012).
\newblock {MuJoCo}: A physics engine for model-based control.
\newblock In \emph{Proc. IROS 2012}, 5026--5033.

\bibitem[{Wachter et~al.(2024)Wachter, Hartl-Nesic, and Kugi}]{wachter2024robot}
Wachter, A., Hartl-Nesic, C., and Kugi, A. (2024).
\newblock Robot base placement optimization for pick-and-place sequences in industrial environments.
\newblock \emph{IFAC-PapersOnLine}, 58(19), 19--24.

\bibitem[{Weingartshofer et~al.(2021)Weingartshofer, Hartl-Nesic, and Kugi}]{weingartshofer2021optimal}
Weingartshofer, T., Hartl-Nesic, C., and Kugi, A. (2021).
\newblock Optimal {TCP} and robot base placement for a set of complex continuous paths.
\newblock In \emph{Proc. ICRA 2021}, 9659--9665.

\bibitem[{Whitman and Choset(2018)}]{whitman2018task}
Whitman, J. and Choset, H. (2018).
\newblock Task-specific manipulator design and trajectory synthesis.
\newblock \emph{IEEE Robotics and Automation Letters}, 4(2), 301--308.

\bibitem[{Zimmermann et~al.(2020)Zimmermann, Hakimifard, Zamora, Poranne, and Coros}]{zimmermann2020multi}
Zimmermann, S., Hakimifard, G., Zamora, M., Poranne, R., and Coros, S. (2020).
\newblock A multi-level optimization framework for simultaneous grasping and motion planning.
\newblock \emph{IEEE Robotics and Automation Letters}, 5(2), 2966--2972.

\end{thebibliography}

\end{document}